\documentclass[letterpaper]{article} 
\usepackage{aaai24}  
\usepackage{times}  
\usepackage{helvet}  
\usepackage{courier}  
\usepackage[hyphens]{url}  
\usepackage{graphicx} 
\usepackage{natbib}  
\usepackage{caption} 
\usepackage{algorithm}
\usepackage{algorithmic}
\usepackage{multirow} 
\usepackage{enumitem}

\usepackage{newfloat}
\usepackage{listings}
\DeclareCaptionStyle{ruled}{labelfont=normalfont,labelsep=colon,strut=off} 
\floatstyle{ruled}
\newfloat{listing}{tb}{lst}{}
\floatname{listing}{Listing}
\title{Cached Transformers:  Improving  Transformers with Differentiable Memory Cache}
\author{
    Zhaoyang Zhang\textsuperscript{\rm 1},
    Wenqi Shao\textsuperscript{\rm 1},
    Yixiao Ge\textsuperscript{\rm 3},
    Xiaogang Wang\textsuperscript{\rm 1},
    Jinwei Gu\textsuperscript{\rm 1},
    Ping Luo\textsuperscript{\rm 2}
}
\affiliations{
    \textsuperscript{\rm 1}The Chinese University of Hong Kong \textsuperscript{\rm 2}The University of Hong Kong 
    \textsuperscript{\rm 3}Tencent Inc\\

}

\usepackage{bibentry}

\begin{document}




\maketitle
\vspace{-10pt}



\begin{abstract}
This work introduces a new Transformer model called Cached Transformer, which uses Gated Recurrent Cached (GRC) attention to extend the self-attention mechanism with a differentiable memory cache of tokens. GRC attention enables attending to both past and current tokens, increasing the receptive field of attention and allowing for exploring long-range dependencies. By utilizing a recurrent gating unit to continuously update the cache, our model achieves significant advancements in \textbf{six} language and vision tasks, including language modeling, machine translation, ListOPs, image classification, object detection, and instance segmentation. Furthermore, our approach surpasses previous memory-based techniques in tasks such as language modeling and displays the ability to be applied to a broader range of situations.

\end{abstract}

\section{Introduction}
The design of Transformer   ~\cite{vaswani2017attention}, a deep model stacking self-attention and feed-forward layers, has achieved remarkable progress in various  tasks.
%
%
Compared to the traditional deep models, a key characteristic  of Transformer  is the self-attention mechanism, which enables global receptive field and allows each token to access all the other tokens in a data batch, providing a flexible scheme to capture  contextual representation ~\cite{vaswani2017attention, dosovitskiy2021an, carion2020end}  .
Such paradigm is however in a complexity square to sequence length, thus not suitable to model long-term dependencies.
In this work, we aim to extend the conventional transformer  models using attention with a long-term token representation  in a memory cache, which enables larger and longer receptive field at minimal additional computations.

%
%

\begin{figure}
\centering
    \includegraphics[width=0.9\linewidth]{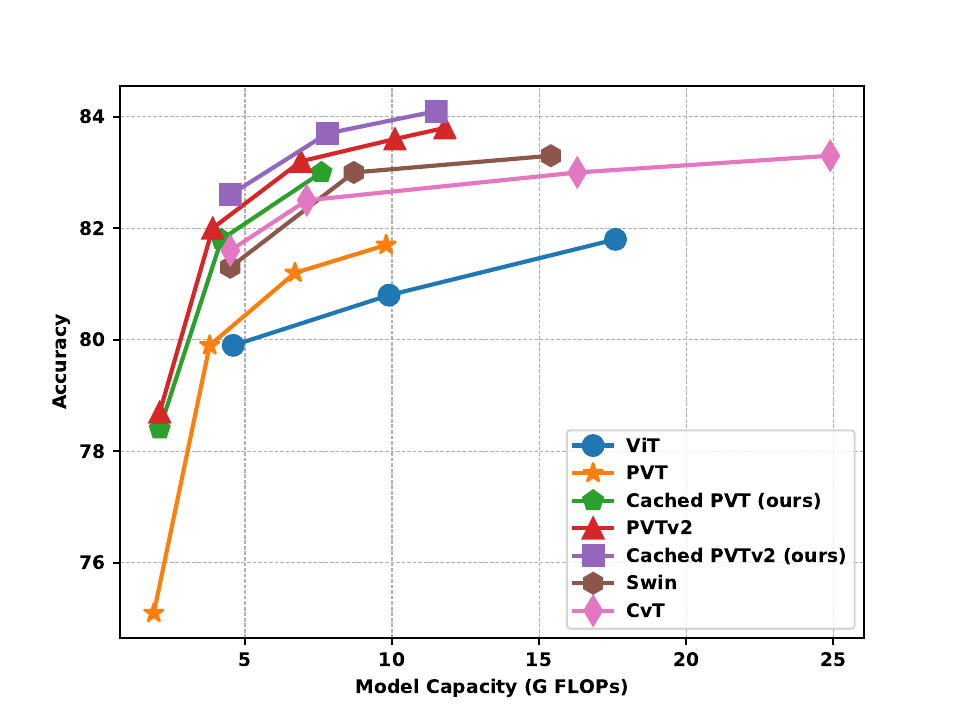}
\caption{\small  \textbf{Performance-Complexity Curve}:  Top-1 accuracy(\%) comparisons on ImageNet with respect to model capacity(G FLOPs) using vision transformers(Parameter-efficiency curves).Curves of our cached models are consistently on top of their corresponding baselines (PVT and PVTv2), indicating the effectiveness of GRC-cached models considering both complexity and accuracy. 
}
\vspace{-20pt}
\label{fig:sota}
\end{figure}

Capturing long-range relationships between tokens and samples is crucial for various tasks due to several reasons.
\label{sec:importanceok}
(i) In sequential data such as language sentences, there can exist dependencies between tokens that are far away from each other. For example, an event or character can be referred to from time to time across multiple paragraphs in an article. Failing to capture such dependencies can result in poor performance in natural language processing tasks.
(ii) Modeling cross-sample relationships can also be useful for non-sequential data like images. 
For example, incorporating a memory module that stores prototypical feature representations can enable instance-invariant feature learning, leading to improved performance in vision tasks ~\cite{long2022retrieval,deng2022fine}. 
Furthermore, other studies ~\cite{wang2020cross, zhong2019invariance} have demonstrated that using cross-batch memory to store previous embeddings can be beneficial for visual representation learning.
(iii) Longer-range attention has also been shown to enhance the representation learning ability of models, as demonstrated in works like ~\cite{dai2019transformer, wu2022memorizing, tay2021omninet}.

However, longer dependency modeling makes computations more expensive. For example, the vanilla Transformer has $O(T^2)$ computational complexity in each attention module when handling a token sequence of length $T$. Although some works apply efficient alternatives, such as low-rank decomposition ~\cite{wang2020linformer, zhu2021long}, block-based sparsification ~\cite{zaheer2020big}, and local sensitive hashing ~\cite{kitaev2020reformer}, they still have complexity linear to the token length ($O(T)$) and thus unable to efficiently capture sparse long-range dependency. Another line of research ~\cite{wu2022memorizing} reduces the complexity of attention module by selecting top-$k$ token pairs from a memory cache for the current tokens, but the cost of maintaining a huge cache of tokens for all layers is still significant. Hence, developing efficient and effective mechanisms for capturing long-range dependencies remains an active area of research.
%


To address these issues, we propose a novel family of Transformer models called Cached Transformer, which has a Gated Recurrent Cache (GRC) that enables Transformers to access historical knowledge, as ilustrated in Fig.~\ref{fig:attn}. The GRC is implemented as a meta-learner that compresses the historical representation into embedding vectors and updates them adaptively with a gating mechanism, avoiding the need for a large memory cache. The GRC updates the past representation with a reset gate that suppresses historical caches and an update gate that further updates the suppressed caches using the current token sequences. This design allows the GRC to access previously seen knowledge in a computationally efficient way. Based on the GRC, we implement a semi-cached attention mechanism that attends to both the latent and current tokens.


We propose Cached Transformer with Gated Recurrent Cache (GRC) and make the following \textbf{contributions}, which make it more appealing than prior arts in several aspects. 
\begin{itemize}[noitemsep]
    \item  GRC is built on a general differentiable formulation and is compatible with various attention schemes, Transformer networks, and tasks. We demonstrate that GRC can be easily plugged into diverse Transformer-variants such as Transformer-XL~\cite{dai2019transformer}, ViT~\cite{dosovitskiy2021an}, PVT~\cite{wang2021pyramid, wang2022pvt}, Swin~\cite{liu2021swin}  Bigbird ~\cite{zaheer2020big}, and Reformer ~\cite{kitaev2020reformer}. 
    \item  GRC can cache all representations of arbitrary length recurrently, independent of sequence length, while existing cache-based methods can only capture recent tokens ~\cite{rae2019compressive, dai2019transformer} or require KNN searching at each step ~\cite{wu2022memorizing}. 
    \item  Besides efficiency, GRC surpasses previous memory-based methods~\cite{dai2019transformer, burtsev2020memory, bulatov2022recurrent} by a large margin on both vision (Table~\ref{tab:attn}) and language tasks (Table~\ref{tab:lm}).
    \item GRC yields consistent improvements not only in sequential data such as texts but also in spatial context such as image classification (Table~\ref{tab:imgnet}) and object detection (Table~\ref{tab:det}). To our knowledge, existing works of Vision Transformers mainly focused on learning intra-sample tokens, while GRC is the first attempt to model cross-sample relationships by attending over inter-sample tokens, such as tokens from different independent images. 
    \item We observe that models with GRC may attend more over the cache than the regular self-attention. We investigate this behavior in image classification and find that GRC can separate features into two parts, attending over caches yielding \textbf{instance-invariant} features, as well as attending over self, yielding \textbf{instance-specific} features (See in Fig.~\ref{fig:pub}). This behavior is similar to that of a vector prototype ~\cite{caron2020unsupervised}, which enables cross-sample regularization to avoid overfitting. 
\end{itemize}

Extensive experiments show that the Cached Transformer  with GRC achieves promising results on various vision and language Transformer backbones. 
(i) \textbf{Language}: In the IWSLT14 De-En benchmark for machine translation, PreNormed Transformer+GRC yields $36.0$ BLEU, outperforming the baselines by $0.5$.
In the challenging long-range-arena benchmark  ~\cite{tay2021long}, GRC improves state-of-the-art methods  with different attention types including Reformer  ~\cite{kitaev2020reformer}, Bigbird  ~\cite{zaheer2020big}, and regular Transformer  ~\cite{vaswani2017attention} consistently by up to $1.2\%$ accuracy.
(ii) \textbf{Vision}: For image classification on ImageNet  ~\cite{krizhevsky2012imagenet}, we plug GRC into the recent vision transformers of different scales, such as ViT  ~\cite{dosovitskiy2021an}, PVT  ~\cite{wang2021pyramid}, PVTv2  ~\cite{wang2022pvt}, Swin   ~\cite{liu2021swin}, and obtain up to $3.3\%$ accuracy gain. 
As shown in Fig.~\ref{fig:sota}, our cached model with PVTv2 backbone achieves superior performance considering both the model \textbf{complexity and accuracy}. 
We further evaluate GRC on the COCO   ~\cite{lin2014microsoft} dataset for object detection and instance segmentation, where PVT+GRC can yield more than $4.0$ box AP improvement.

\section{Related works}

\paragraph{Cached Language Models.}
Cache models are effective in long-range modeling , and are firstly introduced by  ~\cite{kupiec1989probabilistic, kuhn1990cache} for speech recognition. In general, a cache model stores representations of the past, which are usually unigrams or key-value pairs for future computation.  
Transformer-XL ~\cite{dai2019transformer} further applies this technique to transformers, where the cache stores previous key-value pairs in attentions from previous training steps. 
Many memory-based methods are explored following Transformer-XL:
For instance, MT~\cite{burtsev2020memory} and RMT~\cite{bulatov2022recurrent} use extra memory tokens to store local and global information for different segments of inputs.
~\cite{rae2019compressive} compress the tokens before they're saved in the cache to reduce memories and computations. 
However, these methods often use cache in a fixed-length and first-in-first-out (FIFO) manner, which  limits the amount of tokens that can be memorized in sequence. 
%
In contrast, our proposed GRC-based Cached Transformers learn to build the cache adaptively with a complexity that is independent of the attention range. 

\paragraph{Vision Transformers.}
Vision transformers and their variants have recently achieved remarkable success in various vision tasks. The original Vision Transformer (ViT) model ~\cite{dosovitskiy2021an} was the first to split images into patch sequences and feed them into transformer encoders. Despite producing competitive results compared to convolutional neural networks (CNNs), ViTs require costly pretraining on large-scale datasets like JFT-300M  ~\cite{sun2017revisiting}.
To address this issue, several works ~\cite{shao2022dynamic}  attribute it to the lack of inductive bias in ViTs and propose introducing convolutional priors to encode inductive bias such as local context. For example, DeiT  ~\cite{touvron2021going} uses a convolutional teacher to distill knowledge for the transformers, Swin-Transformer ~\cite{liu2021swin} conducts attention in sliding windows, and ConViT  ~\cite{d2021convit}  uses a "soft" convolutional module to encode locality. 
%
However, existing methods focus mainly on intra-sample tokens, whereas our proposed GRC enhances vision transformers by learning instance-invariant features via attending over inter-sample tokens. This allows GRC-based transformers to capture richer contextual information and achieve even better performance on vision tasks.

\section{Methodology}
In this section, we first revisit the vanilla language and vision transformer models, then introduce implementation of Cached Transformers with Gated Recurrent Cache(GRC). 


\subsection{Vanilla Transformer} 
We begin with a brief review of the standard transformer architecture.  The transformer model  ~\cite{vaswani2017attention} is constructed by stacking multi-head self-attention blocks and feed-forward layers which is usually a two-layer linear transformation with activation. Each transformer block is fed with $T\times D$ input tokens, where $T$ is the number of tokens and $D$ represents the size of token embedding.  

\begin{figure}
\centering
    \includegraphics[scale=0.3]{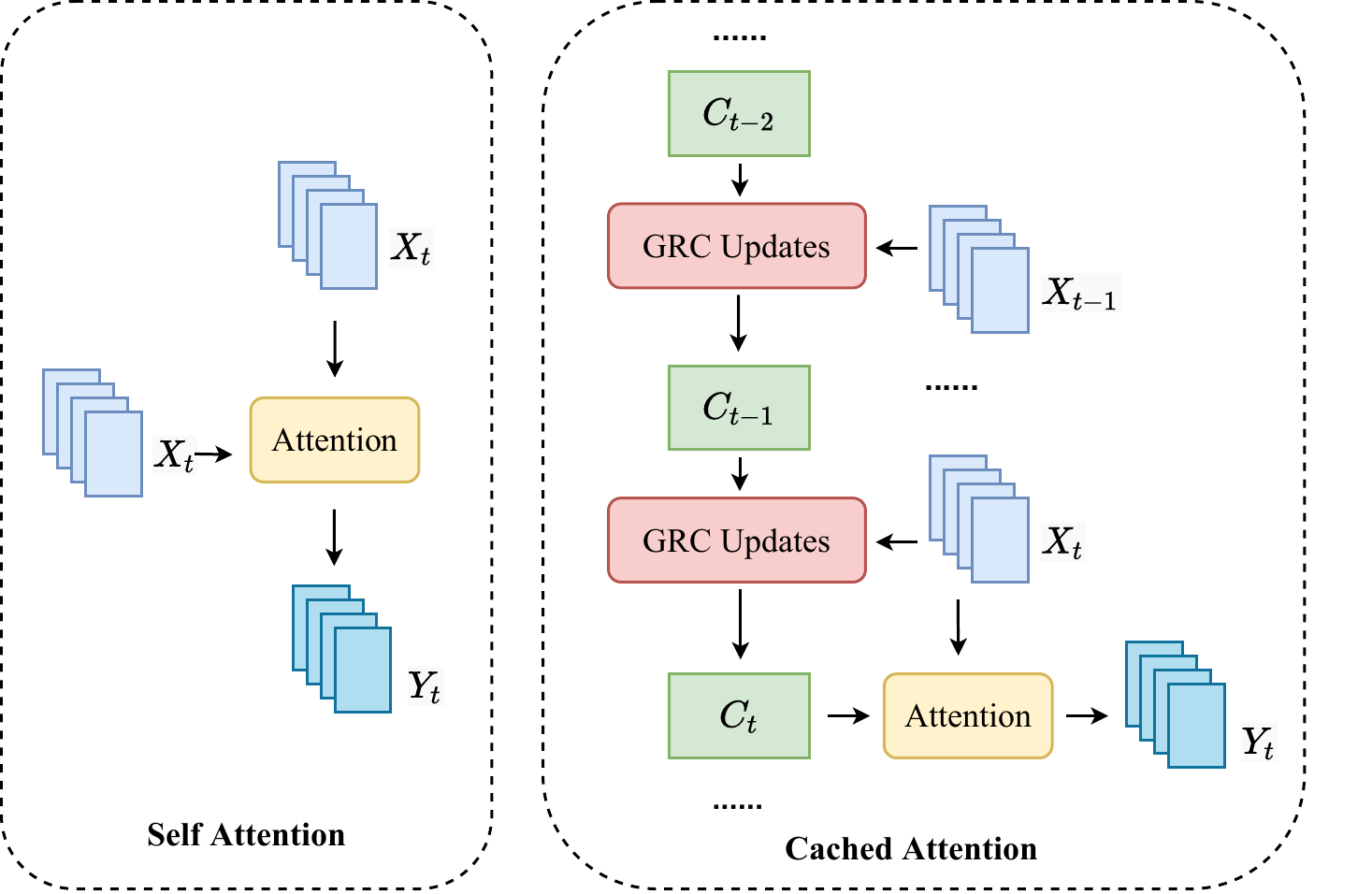}
\caption{\small  Comparisons of vanilla self-attention and cached attentions at training stage. The self-attention only attends to the token itself ($X_t$). 
While in cached attention, outputs at training step $t$ (denoted by $Y_t$) are derived by attending over a Gated Recurrent Cache (GRC, i.e., $C_t$ derived from historical tokens $X_0$ to $X_{t}$), and the current token ($X_t$).
    \label{fig:attn}
    \vspace{-15pt}
}
\end{figure}

\textbf{Self-attention mechanism.} As shown in Fig.\ref{fig:attn}, the self-attention module first projects each input $X$ 
into $Q$ (query), $K$ (key), and $V$(value) using linear transformations. Typically, the self-attention is performed in a multi-head manner where the input will be divided into multiple heads for parallel computation. The output of the attention head $h$ can be written as :
\begin{equation}\label{eq:vanilla-attn}
    o_{self}^{h} = \mathrm{softmax}({Q_{h} K_{h}^T}   /{\sqrt{D/H}}) V_{h},
\end{equation}
where  $o^{h}_{self}$  is the output of head $h$ of the self-attention and $H$ is the number of heads. The output from heads will be concatenated and then fed into another linear transformations with normalization and residual connections.


%

\textbf{Limitations.} 
As shown in Eqn.(\ref{eq:vanilla-attn}),  the vanilla self-attention mechanism used in Transformers is highly sensitive to sequence length, with a computational complexity of $O(T^2)$ with respect to the sequence length $T$. This means that the computational cost grows rapidly as the sequence length increases, which limits the model's ability to capture long-term relationships in the data. As a result, vanilla Transformers can only model relatively short sequences of tokens in language tasks, and it also makes it challenging to develop  cross-task memory modules~\cite{wang2020cross, zhong2019invariance} in a attention-based way for vision tasks.
Towards this issue, we introduce the proposed Cached Transformers, which provides a more flexible paradigm for capturing long-term dependencies, leading to consistent improvements for both vision and language tasks.



\begin{figure*}
\centering
 \small
  \includegraphics[width=0.95\linewidth]{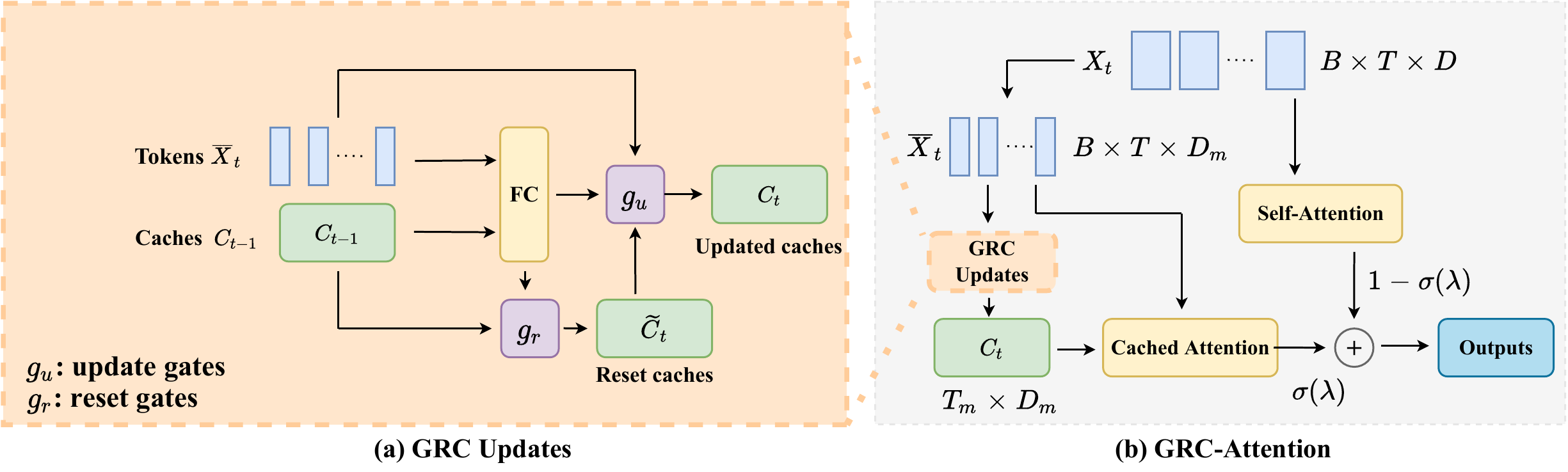}
\caption{ \small The illustration of proposed GRC-Attention in Cached Transformers. (a) Details of the updating process of Gated Recurrent Cache. The updated cache $C_t$ is derived based on  current tokens $\bar{X}_t$ and cache of last step $C_{t-1}$.  The reset gates $g_r$ reset the previous cache $C_{t-1}$ to reset cache $\tilde{C}_{t}$, and the update gates $g_u$ controls the update intensity.  
(b) Overall pipeline of GRC-Attention. Inputs will attend over cache and themselves respectively, and the outputs are formulated as interpolation of the two attention results. 
}
\label{fig:method}
\vspace{-0.15in}
\end{figure*}

\subsection{Cached Transformer}
\label{sec:3.2}
To extend receptive fields of both language and vision transformers, in this section we will introduce our implementations of Cached Transformers, which maintains a continuous cache termed Gated Recurrent Cache (GRC) to support efficient long-term representation learning. 
%
%
The core idea is to hold token embedding as caches which can dynamically record historical samples according to their significance. 
The Cached Transformer will then gain additional capabilities to encode both the current and accumulated information by attending to the gathering of caches $C$ and inputs $X$.
Such an attention scheme is described as GRC-Attention, and the following parts present more details.

\textbf{General implementations.} The proposed Cached Transformers enable attending over caches on arbitrary multi-layers architectures accepting sequential inputs.  
Typically, the Cached Transformer models can be derived by replacing  their self-attention blocks with the proposed GRC-Attention. 
Fig.~\ref{fig:method} (b) gives overall illustrations of how the GRC-Attention is conducted. 

Considering input sequence $X_{t} \in \mathbf{R}^{B\times T \times D}$, where $B$ is the batch size and $t$ denotes training steps, GRC-attention attends to both the memory cache and the current tokens. We formulate GRC-attention by
\begin{equation}\label{eq:GRC-attn}
    O^h = \sigma(\lambda_{h}) * o_{mem}^h + (1 -  \sigma(\lambda_{h}) ) * o_{self}^h ,
\end{equation}
where $O^{h}$ and $o^{h}_{mem}$ are the outputs of the GRC-attention and Cached attention (i.e., attention over memory cache) in the head $h$, respectively. $o^{h}_{self}$ is the output of the self-attention in Eqn.(\ref{eq:vanilla-attn}). Moreover, in Eqn.(\ref{eq:GRC-attn}), $\sigma(\cdot)$ is the sigmoid function and $\lambda_{h}$ is a head-wise learnable ratio trading off self-attention and Cached attention \footnote{All of the $\lambda_{h}$ is initialized to be 0.}.
%

To construct the triplet key, query and value for Cached attention, we choose a portion of $X_t$ as input $\bar{X}_t \in  \mathbf{R}^{B\times T \times D_m}$, which is derived by slicing $X_t$ on channel dimension. Note that  $D_m = r D$\footnote{At most cases we adopt $D_m = \frac{D}{2}$ to reduce the complexity of Cached attention , which means we choose half of the inputs to update caches} indicates channels used for memorizing the past tokens embedding, where $r$ is the caching ratio. With $\bar{X}_t$, the accumulated cache $C_{t-1}$ will then be updated to $C_{t}$ according to the GRC update rules as shown in Fig.~\ref{fig:method}. We describe the construction of GRC in Sec~\ref{sec:sec3.3} in detail.
The Cached attention can be then conducted by using $\bar{X}_t$ as queries and $C_t$ as keys and values, written as:
\begin{equation}
     o_{mem}^{h} =  \mathrm{softmax}({\bar{Q}_{h} \bar{K}_{h}^T}  / {\sqrt{D_m/H}}) \bar{V}_{h}, 
\end{equation}
where $\bar{Q}_h$, $\bar{K}_h$ and  $\bar{V}_h$ are obtained by linear projections of $h$-th head of $\bar{X}_t$, $C_t$ and $C_t$ respectively. 

 \textbf{Generalizations.} Note that while we typically formulate Cached Transformer as a self-attention based model, it can also be an arbitrary transformer variant. In other words, the attention mechanism used to acquire $o^{h}_{self}$ and $o^{h}_{mem}$ in Eqn.(\ref{eq:GRC-attn}) can be substituted by any other attention-like functions, such as sparse attentions ~\cite{zaheer2020big} or local hashing ~\cite{kitaev2020reformer}. Further experiments will provide validations of Cached Transformers on several transformer variants.

\subsection{Gated Recurrent Cache Update}\label{sec:sec3.3}
This section describes the formulation and updating of proposed Gated Recurrent Cache (GRC).   

\textbf{Cache Initialization.} 
The GRC is characterized to be fixed-length vectors $C_{t} \in \mathbf{R}^{T_m \times D_m}$. Unlike previous works that formulate cache to be tokens or words directly ~\cite{tu2018learning, dai2019transformer}, GRC embeds historical tokens implicitly. 
By learning to embed arbitrary length samples into $C_{t}$, GRC allows traversing caches in constant time that is independent of the number of memorized tokens. 
The cache $C_0$ will be initialized to be $T_m$-length zero vectors before training, and then updated as depicted in Fig.~\ref{fig:method}(a). 
%

\textbf{Gating Mechanism.}
Inspired by gated RNNs~\cite{cho2014properties}, we adopt the gating mechanism to enable GRC to dynamically capture dependencies at different time scales.  Specifically, the updating process of $C_t$ is filtered by update gates $g_{u}$ and reset gates $g_{r}$.
Considering updating GRC at time step $t$, we first calculate the  gates $g_{u}$ and $g_{r}$: 
\begin{equation}
    g_{u} = \sigma( W_u [\bar{X}_t, C_{t-1}] )\,\, \mathrm{and}\,\, g_{r} = \sigma( W_r [\bar{X}_t, C_{t-1}] ),
\end{equation}
where $\sigma$ denotes sigmoid function and $[\cdot,\cdot]$ concatenates tokens in channel dimension. For valid concatenation, $\bar{X}_t$ is interpolated into a $T_m$-by-$D_m$ token.
The updated cache $C_{t} $ is formulated by a linear interpolation as given by:
\begin{equation}
    C_{t} = (1 - g_{u})C_{t-1} + g_{u}\tilde{C_{t}}\,\, \mathrm{and}\,\, \tilde{C}_{t} = W_c [\bar{X}_t,  g_{r} \odot C_{t-1}]
\end{equation}
where $\odot$ is element-wise multiplication.  In above process, the  update gates $g_{u}$ decides how much current sample $\bar{X}_{t}$ updates the cache and the reset gates $g_{r}$ suppress the accumulated cache to forget unimportant components.
Note that shape of the derived $C_t$ is $B\times T_m \times D_m$ as $X_t$ is involved, and we therefore average across the batch dimension to fit the cache size. 
%

%



\begin{figure*}[t!]
  \includegraphics[width=0.85\linewidth]{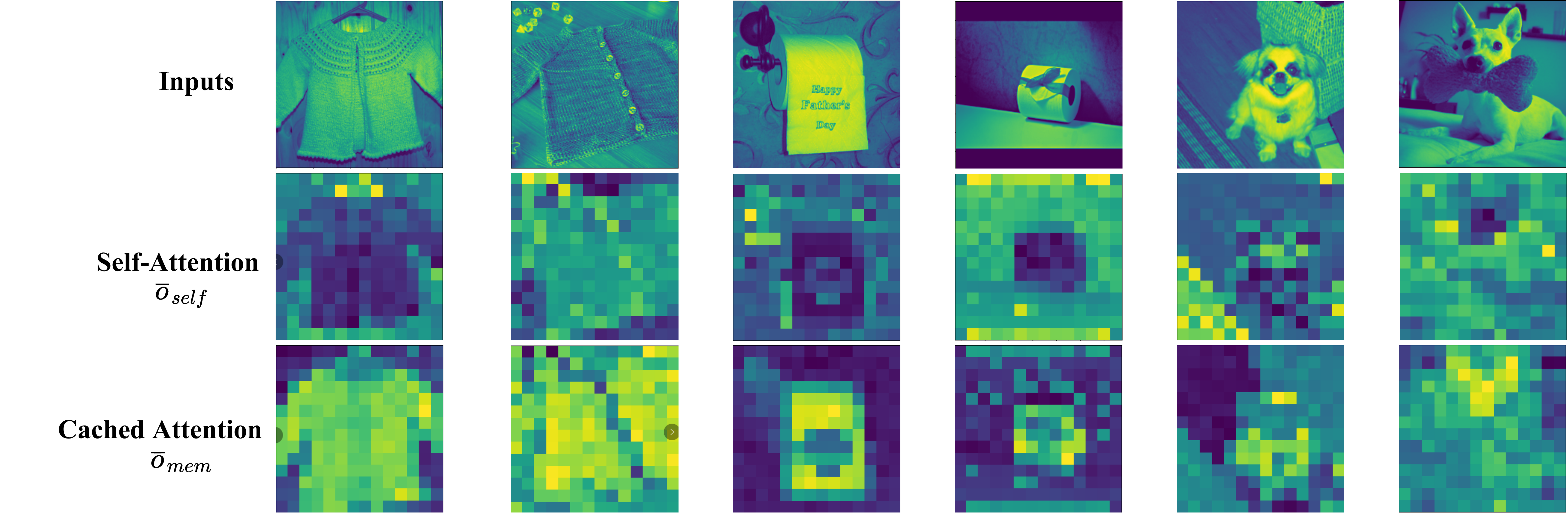}
\caption{
\small{
Visualizations of averaged features output from self-attention and cached attention, which is obtained by feeding images of ImageNet validation sets to trained \textbf{cached ViT-S}.
The results are obtained by averaging features over channel(and head) dimension.
Both  $\bar{o}_{self}$ and $\bar{o}_{mem}$  are unflattened to $14 \times 14$ for better comparisons. Dark pixels mean small values. }
\vspace{-10pt}
}
\label{fig:pub}
\end{figure*}

\section{Experiments}

This section extensively evaluates the effectiveness of the proposed Cached Transformer and Gated Recurrent Cache (GRC) in both vision and language tasks, including language modeling on WikiText-103, Long Listops of Long Range Arena ~\cite{tay2021long}, machine translation on IWSLT14 ~\cite{cettolo-etal-2014-report} / IWSLT15 ~\cite{cettolo-etal-2015-iwslt}, image classification on  ImageNet ~\cite{krizhevsky2012imagenet}, and object detection and instance segmentation on COCO2017 ~\cite{lin2014microsoft}.  In addition, as the cached models are newly introduced to vision transformers, we also perform thorough discussions on the role of the proposed caches and their significance. 
All of the experiments are conducted on Tesla V100 GPUs.

\subsection{Image Classification}
\begin{table}[t!]
\small
        \centering
        \caption{ \small  Performance of various Cached Transformers evaluated on ImageNet.  "(Cached)" indicates models implemented with the proposed GRC-Attention. Top-1 / Top-5 /  $\Delta$ Top-1  denotes top-1 accuracy / top-5 accuracy / top-1 accuracy difference respectively. The cached models outperform their corresponding baselines consistently. }
        \scalebox{0.8}{
        \begin{tabular}{l| c|c|c}
        \hline
        Architecture  & Top-1 (\%) & Top-5 (\%) & $\Delta$ Top-1 (\%)  \\
        \hline
        ViT-S   & 79.9 & 95.0 & - \\
        ViT-S (Cached)  & \textbf{81.3} & 95.5 & + 1.4 \\
        \hline 
        PVT-Tiny  & 75.1 & 92.3 & - \\
        PVT-Tiny (Cached) & \textbf{78.4 }& 94.2 & + 3.3\\
        PVT-Small  & 79.9 & 95.0 & -\\
        PVT-Small (Cached) &  \textbf{81.8} & 95.9 & + 1.9\\
        PVT-Medium & 81.2 & 95.7 & - \\
        PVT-Medium (Cached) & \textbf{83.0} & 96.4 & + 1.8\\
        \hline
        Swin-T & 81.2 & 95.5 & - \\ 
        Swin-T (Cached) & \textbf{82.1} & 95.9 & + 0.9 \\ 
        \hline
        PVTv2-B2 &  82.0 & 95.9 & -\\
        PVTv2-B2 (Cached)  & \textbf{82.6} & 96.2 & + 0.6 \\
        PVTv2-B& 83.2 & 96.3 & - \\
        PVTv2-B3 (Cached)  & \textbf{83.7} & 96.4 & + 0.5  \\
        PVTv2-B4 & 83.6 & 96.3 & - \\
        PVTv2-B4 (Cached ) &\textbf{84.1} & 96.6 & + 0.5 \\
        \hline
    
        \end{tabular}}
        \vspace{-10pt}
        \label{tab:imgnet}
\end{table}

\paragraph{Experiments Setup.} We first evaluate our methods on Imagenet-1k for image classification.
We implement our GRC-Attention as a general pytorch module which maintains fixed-length buffers as cache.
In image classification task, we set the cache ratio $r$ to be $0.5$ and keep cache length $T_m$ equal to the length of image patches $T$.
For fair comparisons, we directly replace the self-attention layers in corresponding transformers with our GRC-Attention module without varying the architecture and hyperparameters. 
To maintain spatial token structures, we add positional encodings to our proposed GRC-Attention like other vision transformers.
Both the baselines and their cached counterparts are trained with $224 \times 224$ size inputs using 16 GPUs.
To fully validate the proposed cache mechanism, we evaluate GRC-Attention on four recent vision transformers including: ViTs ~\cite{dosovitskiy2021an}, PVT ~\cite{wang2021pyramid}, Swin-Transformer ~\cite{liu2021swin} and PVT-v2 ~\cite{wang2022pvt}. 
Without bells and whistles, all of the training settings for cached models are kept consistent with the original baselines including data augmentation, optimizer type, learning rates and training epochs. 
\begin{figure*}
 \small
\centering
  \includegraphics[width=0.99\linewidth]{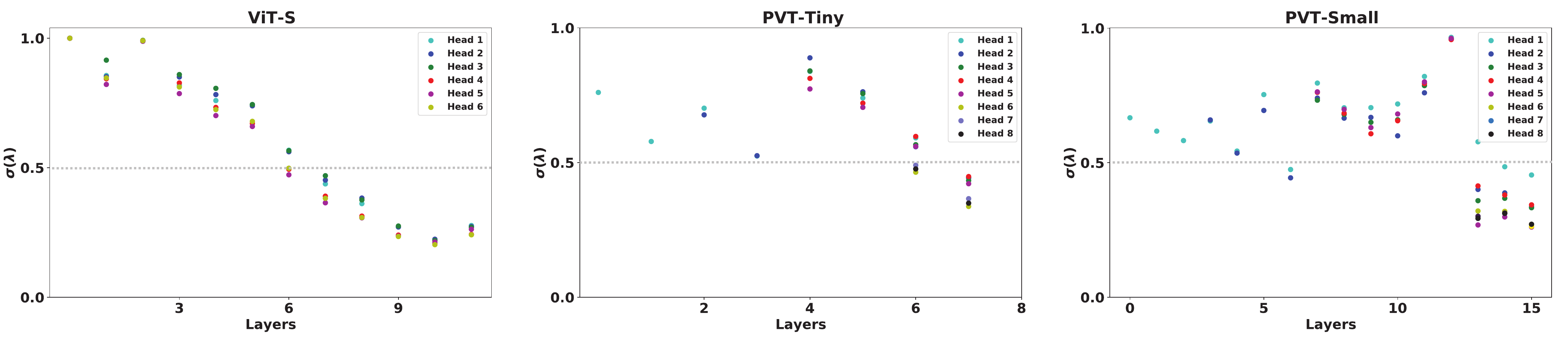}
\caption{\small{Visualizations of learned $\sigma(\lambda^h)$ for each head with respect to layer number (from shallow to deep) in different models: ViT-S, PVT-Tiny and PVT-Small. Note that the ViT-S has 6 heads for all the layers, while PVT-Tiny and PVT-Small adopt a progressive head strategy where head numbers increase from 1 to 8 gradually.
Circles with different colors denote those different heads.   $\sigma(\lambda^h)$ controls the interpolation ratio of cached attention outputs $o_{mem}$ which reflects head-wise contributions of cached attention to the final outputs. Note that $\sigma(\lambda^h) > 0.5$ means cached attention contributes more than self-attention. As shown, in all of the three models,  $\sigma(\lambda^h) > 0.5$ holds for more than half of the GRC-Attention layers, deducing that the model outputs are significantly dependent on the cache. 
%
}}
\label{fig:lam}
\vspace{-12pt}

\end{figure*}

%

\paragraph{Classification Results.}
Table~\ref{tab:imgnet} reports overall performance of cached transformers on corresponding baselines. As shown, transformers implemented with GRC-Attention consistently outperform their no-cache counterparts by yielding significantly higher accuracy, demonstrating the effectiveness of our proposed caching mechanism. 
For instance, by enabling cache, PVT-Tiny can achieve  $78.4\%$  top-1 accuracy and $94.2\%$ top-5 accuracy, surpassing the original PVT-Tiny by  $3.3\%$ and $1.9\%$ respectively. 
%
%
Moreover, even for the recent stronger backbone PVTv2, our proposed cached mechanism can still keep $>0.5$ top-1 improvements. 

\paragraph{Complexity Analysis.}
 In current settings where cache ratio $r=0.5$, replacing all the attention layers with GRC-Attention will cost approximately an extra $10\% - 15\%$  FLOPs and Params. 
Considering the performance improvements, the extra computations are acceptable (See in Fig.~\ref{fig:sota}) and more efficient than increasing the depth and width of models. 
%
%

\paragraph{Significance of Cached Attention.}
To verify that the above performance gains mainly come from attending over caches, we analyze the contribution of $o_{mem}$ by visualizing the learnable attention ratio $\sigma(\lambda^h)$. 
Please be reminded that in Eq~\ref{eq:GRC-attn}, outputs of GRC-Attention is derived by interpolating outputs of cached attention $o_{mem}^h$ and self-attention $o_{self}^h$  according to $\sigma(\lambda^h)$.
Hence,  $\sigma(\lambda^h)$ can be used to represent the relative significance of $o_{mem}^h$ and $o_{self}^h$.
Fig.~\ref{fig:lam} depicts the learned $\sigma(\lambda^h)$ for each head respect to layers in ViT-S, PVT-Tiny and PVT-Small. 
As we can see, for more than half of the layers,  $\sigma(\lambda^h)$ is larger than $0.5$, denoting that outputs of those layers are highly dependent on the cached attention. 
Besides, we also notice an interesting fact that the models always prefer more cached attention except for the last several layers.
This makes us curious about the roles of cached attention: what is the feature that models actually learn by attending over caches?
The following paragraph answers this question.

\paragraph{Roles of Cached Attention.} We investigate the function of GRC-Attention by visualizing their interior feature maps. We choose the middle layers of cached ViT-S, averaging the  outputs from self-attention $o_{self}$ and cached attention ($o_{mem}$) across the head and channel dimension, and then normalizing them into $[0, 1]$. The corresponding results are denoting as $\bar{o}_{self}$ and $\bar{o}_{mem}$, respectively. 
Fig.~\ref{fig:pub} provides visualizations of $\bar{o}_{self}$ and $\bar{o}_{mem}$ obtained by feedings images of ImageNet validation sets to trained cached ViT-S. 
As $\bar{o}_{self}$ and $\bar{o}_{mem}$  are sequences of patches,  they are unflattened to $14 \times 14$ shape for better comparison.
From Fig.~\ref{fig:pub} we can see,  features derived by the above two attentions are visually complementary.  
In GRC-Attention, $o_{mem}$ is derived by attending over the proposed cache (GRC) containing compressive representations of historical samples, and thus being adept in recognizing \textbf{public} and frequently showing-up patches of this \textbf{class}. 
While for $o_{self}$ from self-attention branch, it can focus on finding out more private and \textbf{characteristic} features of current \textbf{instance}. 

With above postulates, we can attempt to explain the regularity of $\sigma(\lambda^h)$ in Fig.~\ref{fig:lam}: employing more $o_{mem}$ (larger $\sigma(\lambda^h)$ ) in former layers can help the network to distinguish this instance coarsely, and employing more $o_{self}$ (smaller $\sigma(\lambda^h)$) enable the model to make fine-grained decision. 

\paragraph{Cross-sample regularization.}
The above paragraph also shows that our proposed cache performs similarly to vector prototypes  ~\cite{caron2020unsupervised},  storing public features of the same class implicitly and allowing models to classify inputs with both the public and characteristic representations.
In such a way, the predictions are not only dependent on the current inputs but also on related cached samples, thus providing a cross-sample regularization to avoid overfitting.


\begin{table}
        \centering
        \caption{\small Performance(Top-1 Accuracy) comparisons of cached models using GRC and attention-based }
                \vspace{-5pt}

        \scalebox{0.9}{
        \begin{tabular}{l | c c c }
                \hline
        Model & 	No cache & 	Attention-based cache &  GRC \\
        \hline
        ViT-S &	79.9 &	80.0  &	\textbf{81.3} \\
        PVT-Tiny  &	75.1 & 74.8 & \textbf{78.4} \\
        PVT-Small & 79.9 & 79.6 &  \textbf{81.8} \\
        \hline
        \end{tabular}}
        \label{tab:attn}
        \vspace{-10pt}
\end{table}

\subsubsection{GRC v.s. other memory-based methods.}
We perform further ablations to compare GRC and attention-based memory for image classification in ImageNet-1k. We deploy Transformer-XL-style caches to Vision Transformers(including ViT-S, PVT-Tiny and PVT-Small) and compare them to corresponding GRC-cached models. As shown in Table~\ref{tab:attn}, GRC-cached models consistently outperform their attention-based cache and no-cache counterparts.
Besides, it can be noted that the attention-based cache can hardly improve the model performance. 



\begin{table}
\small
        \centering  
        \caption{\small Object detection and instance segmentation performance on COCO \textit{val2017} following Mask R-CNN $1\times$ settings. }
        \vspace{-10pt}
        \scalebox{0.81}{
        \begin{tabular}{l |l c c|l c c}
        \hline
        Architecture & AP$^b$ & AP$^b_{50}$  & AP$^b_{75}$ &AP$^{m}$ & AP$^m_{50}$ &  AP$^m_{75}$   \\
        \hline
        PVT-Tiny & 36.7 & 59.2 & 39.3 & 35.1 & 56.7 & 37.3 \\
        + Cached & \textbf{41.0} (+ 4.6) & 63.4 & 44.8 & \textbf{38.3} (+ 3.2) & 60.4 & 41.1
        \\
        \hline
        PVT-Small & 40.4 & 62.9 & 43.8 & 36.3 & 60.1 & 40.3 \\ 
        + Cached & \textbf{44.5} (+ 4.1) & 67.1 & 48.6 & \textbf{41.0} (+ 4.7) & 64.0 & 44.1 \\
        \hline
        PVT-Medium & 42.0 & 64.4 & 45.6 & 39.0 & 61.6 & 42.1 \\
        + Cached  &\textbf{46.6} (+ 4.6) & 68.2 & 51.0 & \textbf{42.3} (+ 3.3)& 65.3 & 45.5 \\
        \hline
        \end{tabular}}
                \vspace{-10pt}

        \label{tab:det}
\end{table}
\begin{table*}[ht!]
        \centering
        \caption{\small Neural machine translation results using Pre-Norm Transformers in terms of BLEU scores. }
        \scalebox{0.85}{
        \begin{tabular}{l | c c c | c c c}
                \hline
        \multirow{2}{*}{Architecture} & \multicolumn{3}{c|}{IWSLT14}  & \multicolumn{3}{c}{IWSLT15} \\
         &  De-En & Es-En & En-Fr & De-En & En-Vi & Cs-En  \\  \hline
        Transformer & 35.5 &  41.4 & 41.5 & 36.1 & 29.8 & 28.8  \\
        Transformer (GRC-cached) & 36.0(+ 0.5) & 41.8(+ 0.4) & 41.7(+ 0.2) & 36.3(+ 0.2) & 30.2(+ 0.4) & 29.4(+ 0.6)  \\
        \hline
        \end{tabular}}
        \label{tab:mt}
\end{table*}

\subsection{Object Detection and Instance Segmentation.}

\paragraph{Experiments Setup.} We further assess the generalization of our GRC-Attention on object detection / instance segmentation track using COCO2017 dataset ~\cite{lin2014microsoft}. The models are trained on the COCO \textit{train2017} (118k images) and evaluated on \textit{val2017} (5k images).
We use the cached PVT as backbone and adopt the Mask R-CNN detector ~\cite{he2017mask} to verify the effectiveness of GRC-Attention. The standard COCO metrics of Average Precision (AP)  for bounding box detection (APbb) and instance segmentation (APm) are used to evaluate our methods. 
All of the training settings and hyperparameters are kept the same as PVT original implementation ~\cite{wang2021pyramid}, and all of the involved models  are trained for 12 epochs  using 8 GPUs.
For both the cached PVT and baselines, backbones are firstly pretrained on ImageNet and then fine-tuned for detection. 

\paragraph{Resuts.} 
As shown in Table~\ref{tab:det}, when using Mask R-CNN for object detection, the cached PVTs significantly outperform their baselines. 
For example, the AP of cached PVT-Medium is 4.6 (46.6 vs.\ 42.0) points better than its no-cache counterparts. Similar results can also be found in instance segmentation results, where cached PVT-Medium achieves 3.3 higher AP$^m$ (39.0 vs.\ 42.3). 
These results demonstrate the generalizability of the proposed caching  mechanism.

\subsection{Language Modeling}

\begin{table}
 \small
 \vspace{-10pt}
\centering
        \caption{
        \small
         Comparison of performance(Test PPL) for GRC and other Memory-based methods~\cite{burtsev2020memory, bulatov2022recurrent} on WikiText-103. The smaller is better.  GRC outperform Transformer-XL and previous memory-based methods for language modeling by a large margin of 1.1 PPL.  }
         \scalebox{0.75}{
        \begin{tabular}{l |c c c c }
        \hline
        Architecture & baseline & MT-cached & RMT-cached & GRC-cached \\
        \hline
        Transformer-XL$_{base}$ & 24.0 & 23.99 & 23.95 & \textbf{22.9} \\
        Transformer-XL$_{large}$ & 18.3 & - & - & \textbf{17.9} \\
        \hline
        \end{tabular}
        }
        \label{tab:lm}
        \vspace{-10pt}
\end{table}

\paragraph{Experimental Setup}
In this work, we conduct experiments to compare the performance of Gated Recurrent Cache (GRC) with Transformer-XL~\cite{dai2019transformer} on a language modeling task using the WikiText-103 benchmark. To implement GRC-cached language models, we use the publicly available fairseq framework and follow the default memory-based Transformer-XL configurations as our baselines, including model architecture and training settings. To ensure a fair comparison, we compare GRC-cached models with two other memory-based methods, Memory Transfomer (MT) ~\cite{burtsev2020memory} and Recurrent Memory Transformer (RMT) ~\cite{bulatov2022recurrent}. We implement GRC-cached models by replacing the caching scheme with the GRC approach while keeping all data augmentation and hyper-parameters unchanged for a more fair comparison.

\paragraph{Comparison to Other Memory-Based Methods}
We present the performance of GRC-cached models compared to Transformer-XL baselines and other memory-based methods in Table~\ref{tab:lm}. The results show that GRC-cached models outperform Transformer-XL and other memory-based methods in terms of perplexity on both base and large-scale models. For instance, GRC-cached Transformer-XL$_{base}$ achieves up to 1.1 lower PPL compared to the baseline Transformer-XL and 1.05 lower PPL to the RMT, demonstrating the superiority of GRC over previous memory-based Transformer methods.



\subsection{Long Range Arena}

\textbf{Experiments Setup.}
We extensively conduct experiments on recently proposed Long Range Arena (LRA) benchmarks ~\cite{tay2021long} to validate our proposed methods under the long-context scenario. 
To demonstrate the long-range sequence modeling capability of GRC-Attention and the corresponding cache mechanism, we choose the challenging Long ListOps task in LRA, which is a longer variation of ListOps task ~\cite{nangia2018listops} with up to 2k length sequences and considerablely difficult. 
In this task, we also extend GRC-Attention to efficient attention variants by replacing the self-attention function (See section \ref{sec:3.2}).
Concretely, we compare GRC-Attention to their no-cache counterparts on baselines including Transformer ~\cite{vaswani2017attention}, BigBird ~\cite{zaheer2020big} and Reformer ~\cite{kitaev2020reformer}.
For those efficient attentions like BigBird and Reformer, we only import gated recurrent cache and maintain their inner attention function unchanged. 
All of the experiments are  under default settings in  ~\cite{tay2021long}.

\paragraph{Results.}
Table~\ref{tab:lra} reports Long ListOps results. As shown, cached models consistently outperform their baselines (including the SOTA methods Reformer) significantly. For instance, by employing GRC, BigBird model can achieve 1.39 higher accuracy. These results show the long-range sequence modeling ability of GRC as well as its generalizability to other attention variants.

\begin{table}
 \small
\centering
        \vspace{-10pt}

        \caption{
        \small
        Results on Long ListOPs task in LRA in terms of accuracy. 
        The "cached" column indicates cached models whose attention layers are implemented as generalized GRC-Attention.  
        $\Delta$ denotes the difference between proposed cached models and baselines.  }
                \vspace{-5pt}

        \begin{tabular}{l |c c c }
        \hline
        Architecture & baseline & GRC-cached & $\Delta$ \\
        \hline
        Transformer & 36.23 &  37.40 & + 1.17 \\
        BigBird & 36.06 & 37.45 & + 1.39 \\
        Reformer & 37.27 & 37.85 & + 0.58 \\
        \hline
        \end{tabular}
        \label{tab:lra}
        \vspace{-10pt}
\end{table}

\subsection{Neural Machine Translation}

\paragraph{Experiments Setups.}
We experiment our methods on widely used public datasets IWSLT14 and IWSLT15.  Multiple language sources\footnote{IWSLT14: German-English(De-En), Spanish-English(Es-En) and  English-French(En-Fr), IWSLT15: German-English(De-En), English-Vietnamese(En-Vi) and Czech-English(Cs-En)}are included to fully verify effectiveness of the proposed GRC, and models are trained for each track individually.
We adopt the Pre-Norm Transformer settings in ~\cite{wang2019learning} and implement the models using \textit{fairseq-py} ~\cite{ott2019fairseq} framework.
Following ~\cite{wang2019learning, ott2019fairseq}, we generally increase the learning rates by 2 and average the last  10 checkpoints for inference. 
We employ the proposed GRC-cached models by replacing all attention modules of transformer encoder layers with GRC-Attention. The cache length $T_m$ is set to be 64 for all cached models. 
All the transformers in this task are using six encoder layers and six decoder layers.
For a fair comparison, both the baselines and cached models are trained under identical settings.

\paragraph{Results.} 
We use BLEU ~\cite{Papineni2002BleuAM} as evaluation metrics and compare GRC cached transformers to their baselines in Table~\ref{tab:mt}. 
%
It can be seen that consistent improvements can be reached by applying GRC-Attention to baselines.
For tracks like IWSLT14 De-En and IWSLT15 Cs-En,  the increments can achieve 0.5/0.6 points, which is actually significant for these tasks.

\section{Discussion}
We introduce Cached Transformer with Gated Recurrent Cache (GRC), a simple extension to Transformer-based models that significantly increases the length of attention context by allowing access to historical states through a gating mechanism.
GRC embeds previous tokens, whether they are close or distant, as fixed-length vectors, without complexity dependence on the number of cached tokens. Consequently, GRC model token dependencies over a broader range of input, resulting in improved accuracy and performance across diverse Transformers-variants with different architectures and attention functions, on a variety of vision and language tasks.





\clearpage
{\small
\bibliography{references.bib}
}

\newpage

\appendix


\section{Full Literature Reviews}

\textbf{Language Transformers.}
Transformer is firstly introduced in language processing by (\cite{vaswani2017attention}), and a great deal of work has been done to improve it. 
The most influential works of language transformers are GPT \cite{radford2018improving, radford2019language, brown2020language} and  BERT \cite{devlin2018bert}. The GPT/BERT family works in a `pretraining-finetuning' fashion, achieving state-of-art performance on various language benchmarks. But they are also expensive on both hardware and energy. 
Another series of methods improve the vanilla transformers and also seek for trade-off between performance and efficiency.  
To cover long-range information in language, several efficient attention techniques are proposed such as kernel approximation \cite{wang2020linformer, choromanski2020rethinking},  sparsification \cite{zaheer2020big, roy2021efficient} and local hashing  \cite{kitaev2020reformer} ). 
Rather than these lightweight attentions, other works attempt to apply attention selectively with predefined rules like sliding windows  \cite{beltagy2020longformer}, hierarchical architecture \cite{ainslie2020etc} and token pruning \cite{sukhbaatar2021not}.

\textbf{Cache Language Models and Memory-based Methods.}
Cache models are effective in long-range modeling , and are firstly introduced by  ~\cite{kupiec1989probabilistic, kuhn1990cache} for speech recognition. In general, a cache model stores representations of the past, which are usually unigrams or key-value pairs for future computation.  ~\cite{grave2016improving} extend such kinds of methods to neural language models(RNNs), where cache stores the most-recent pairs of inputs and hidden states, and  ~\cite{brahma2018improved} then improves neural cache models by decoding the past tokens as a regularization.   Transformer-XL ~\cite{dai2019transformer} further applies this technique to transformers, where the cache stores previous key-value pairs in attentions from previous training steps. 

Many memory-based methods are explored following Transformer-XL:
For instance, MT~\cite{burtsev2020memory} and RMT~\cite{bulatov2022recurrent} use extra memory tokens to store local and global information for different segments of inputs.
~\cite{rae2019compressive} compress the tokens before they're saved in the cache to reduce memories and computations. 
In addition to general representations, some works also store task-specific information in cache to improve performance. For instance, ~\cite{tu2018learning} proposes to enhance neural machine translation models by remembering translation history. However, these methods often use cache in a fixed-length and first-in-first-out (FIFO) manner, which  limits the amount of tokens that can be memorized in sequence. 

To address this issue, recent memory-based work ~\cite{khandelwal2019generalization, wu2022memorizing} proposes to store key-value pairs in a large cache without compression and perform K-nearest neighbor (KNN) lookup to search over them. While this approach yields competitive results in language modeling, it still requires a large memory footprint and significant time for searching, particularly for longer attention ranges. In contrast, our proposed GRC-based Cached Transformers learn to build the cache adaptively with a complexity that is independent of the attention range. 
%

\textbf{Vision Transformers.}
Vision transformers (and variants) have achieved great success in various vision tasks recently. 
ViTs \cite{dosovitskiy2021an} firstly propose to split images into patch sequences and feed them into transformer encoders. 
Although yielding competitive results to CNNs, ViTs have an issue requiring costly pretraining on large-scale datasets like JFT-300M \cite{sun2017revisiting}.
Many works \cite{shao2022dynamic} ascribe this to the lack of inductive bias and propose to introduce convolutional priors to ViTs to encode inductive bias like local context. For example, DeiT \cite{touvron2021going} use a convolution teachers to distill knowledge for the transformers, Swin-Transformer \cite{liu2021swin} conducts attention in sliding windows, and ConViT \cite{d2021convit} use a "soft" convolutional module to encode locality. 
Moreover, other methods like PVT \cite{wang2022pvt}, T2T \cite{yuan2021tokens}, and CVT \cite{wu2021cvt} further improve vision transformers by importing convolutional priors in CNNs \cite{he2016deep}.
Different from existing methods that focus on intra-sample tokens,  the proposed GRC further enhances vision transformers
by modeling dependencies of inter-sample tokens.

\section{Implementation Details}

\subsection{Training and Inference Algorithms}
Algorithm~\ref{alg:grc-attn} provides detailed produces of the proposed GRC-Attention in a forward pass. During training, each GRC-Attention module maintains a continuous cache $C_t$, which will be updated at each iteration. Note that all the computations involved in GRC-Attention are differentiable and corresponding parameters can thus be optimized using gradient-based methods.  
The accumulated caches $C_t$ are stored with network parameters after training, and will be used directly for inference \textbf{without} any further updating.

\subsection{Other implementation details.} 
GRC supports end-to-end training with other network parameters using gradient-based methods. The cache $C_{t}$ will be saved as buffers together with trained models and used for evaluation. Similar to training statistics in batch normalization,  $C_{t}$ will be freezed at inference time and thus require no updating. 
We also notice that  some tasks accept varying length inputs (varying $T$, like object detection or machine translation).
In such cases, we will interpolate valid input sequences(without zero-padding) to the fixed cache length $T_m$ and then continue cache updating.


\begin{algorithm}[tb]
	\caption{Forward pass of GRC-Attention at training stage.}
	\label{alg:grc-attn}
	{\fontsize{9}{9} \selectfont
		\begin{algorithmic}[1]
       \REQUIRE training step $t$ ($t>0$) , mini batch inputs \scalebox{0.9}{ $X \in \mathbf{R}^{B \times T\times D}$}, learnable parameters $\lambda^h$ for head $h \in \{0, 1, ..., H-1\}$, accumulate cache $C_{t-1} \in \mathbf{R}^{ T_m \times D_m}$, where $D_m = r D$ and $r$ is caching ratio. 
        \ENSURE initialize $C_0$ to be zero vectors, $\lambda^h = 0$ for all heads,  caching ratio $r=0.5$, and let $T_m = T$ (for image classification / Long ListOps / Object Detection) or $T_m = 64$ (for Machine Translation). 
        \renewcommand{\algorithmicrequire}
        {\textbf{Output:}}
        
        \REQUIRE the attention outputs $O^h$ over both caches and inputs.
        \STATE calculate $\bar{X}_t \in \mathbf{R}^{B \times T \times D_m}$  by slicing inputs $X_t$ with ratio $r$.
        \STATE interpolating $\bar{X}_t$ to length $T_m$ if $T \neq T_m$.
        \STATE calculate update gates $g_u$ and reset gates $g_r$ following Eqn.(4).
        \STATE calculate $C_t$ following Eqn.(5) and averaging $C_t$ on batch dimension.
        \STATE update $C_{t-1} \longleftarrow C_t$ and store it.
        \STATE calculate self-attention outputs $o_{self}^h$ following Eqn.(1).
        \STATE calculate cached attention outputs $o_{mem}^h$ following Eqn.(3).
        \STATE calculate $O^h$ following Eqn.(2).
        \end{algorithmic}
	}
\end{algorithm}

\begin{table*}[htp!]
        \centering
        
        \caption{ Object detection  performance on COCO \textit{val2017} following RetinaNet $1\times$ settings. }
        \begin{tabular}{l |l c c|l c c}
        \hline
        Architecture & AP & AP$_{50}$  & AP$_{75}$ &AP$_{S}$ & AP$_{M}$ &  AP$_{L}$   \\
        \hline
        PVT-Tiny & 36.7 & 56.9 & 38.9 & 22.6 & 38.8 & 50.0 \\
        PVT-Tiny (Cached) &\textbf{40.2 }(+ 3.5) & 61.1 & 43.1 & 25.0 & 43.7 & 53.4  
        \\
        \hline
        PVT-Small & 40.4 & 61.3 & 43.0 & 25.0 & 42.9 & 55.7 \\
        PVT-Small (Cached) & \textbf{44.0} (+ 3.6) & 65.4 & 47.4 & 29.7 & 47.7 & 57.5 \\
        \hline
        PVT-Medium & 41.9 & 63.1 & 44.3 & 25.0 & 44.9 & 57.6 \\
        PVT-Medium (Cached)  & \textbf{45.7} (+ 3.8) & 67.1 & 49.1 & 29.0  & 49.3  & 60.2  \\
        \hline

        \end{tabular}
        \label{tab:retina}
\end{table*}

\subsection{Experimental details}

\textbf{Image classification on ImageNet.} We evaluate the performance of GRC-Attention using various vision transformers((including ViTs, PVT, Swin, and PVTv2)) on ImageNet-1k\cite{krizhevsky2012imagenet}, which consists of 1.28M training images and 50K validation validation images from 1K classes. 
For each baseline, we implement their cached variants by replacing all of their self-attention layers with GRC-Attention directly, keeping their architectures unchanged.
By default, the cache ratio is set as 0.5 and cache length equals to the patch numbers $T_m = T$.
As suggested by \cite{ramachandran2019stand}, positional encodings are added to GRC-Attentions. 
To fairly compare cached models to their baselines, we adopt their original training settings including  data augmentations, optimizers and other hyperparameters. 
Specifically, we use Adam optimizer with a momentum of 0.9 and a weight decay of 0.05. All of the models are trained in $224\times224$ images for 300 epochs, with cosine learning rate scheduler. 
Both the baselines and cached models use standard timm augmentation like \cite{touvron2021training}, including normalization, random cropping, horizontal flipping and color jittering. 
Global average poolings are used in PVT and Swin, where pooling sizes for the first two blocks are  4 and 2, respectively. 
All of the models are trained on 16 Nvidia Tesla V100 GPUs, with 32 GB memory. 

\paragraph{Object detection and instance segmenation on COCO 2017.}  
The models are trained on the COCO \textit{train2017} (118K images) and evaluated on \textit{val2017} (5K images).
We use the cached PVT as backbone and adopt the Mask R-CNN detector \cite{he2017mask} to verify the effectiveness of GRC-Attention.
Before training, we use the weight pre-trained on ImageNet(from prior experiments) to initialize the backbone except for the cache $C$, which will be initialized to zeros.
As input length ($T$) varies in object detection, at the training stage $\bar{X}$ will be interpolated to be of length $T_m$ to update the cache. 
The standard COCO metrics of Average Precision (AP)  for bounding box detection (APbb) and instance segmentation (APm) are used to evaluate our methods. 
All of the training settings and hyperparameters are kept the same as PVT original implementation \cite{wang2021pyramid}, and all of the involved models are trained for 12 epochs ($1\times$ training schedule) using 8 V100 GPUs.
AdamW\cite{loshchilov2018decoupled} optimizer is adopted with $1\times10^{-4}$ initial learning rates. 
The training images are resized to $800\times1333$, which means the shorter side is 800 pixels and the longer side does not exceed 1333 pixels. 
At the testing stage, the shorter side of the input images is fixed to 800.
For both the cached PVT and baselines, backbones are firstly pretrained on ImageNet and then fine-tuned for object detection.

\begin{table}[h!]
        \centering
                \caption{ Performance of GRC on ImageNet-22k.}
        \begin{tabular}{l |c c  }
        \hline
        Model & Top-1 Acc &  Top-5 Acc   \\ \hline
        Swin-T & 80.9 & 96.0 \\
        Swin-T (Cached) & \textbf{81.7 (+0.8)} & \textbf{96.4 (+0.4)} \\
        \hline
        \end{tabular} 
        \label{tab:im22}
\end{table}

\begin{table*}[ht!]
        \centering
        
        \caption{ Training/inference time for GRC-cached models on ImageNet.  }
        \begin{tabular}{l |c c c | c}
        \hline
        Model & Training throughput & Testing throughput &  FLOPs & Top-1 Accuracy   \\
        \hline
 PVT-Tiny & 313 & 930  & 1.90G & 75.1 \\
 PVT-Tiny(Cached) & 257& 768 & 2.15G & 78.4  \\ \hline
 PVT-Small & 181 & 689 & 3.80G & 79.9 \\
 PVT-Small(Cached) & 146  & 561 & 4.29G & 81.8  \\ \hline
 PVT-Medium & 101 & 393 & 6.70G & 81.2 \\
 PVT-Medium(Cached) & 84 & 319 & 7.61G & 83.0  \\
        \hline
        \end{tabular}
        \label{tab:th}
\end{table*}

\begin{table}[ht!]
        \centering
        
        \caption{ Performance of GRC with different caching ratios. }
        \begin{tabular}{l  c c| c}
        \hline
 Model & Ratio & FLOPs &  Acc \\  
 \hline
 PVT-Tiny & 0.000 & 1.90G & 75.1 \\
 PVT-Tiny & 0.125 & 1.93G & 75.7 \\
 PVT-Tiny & 0.250 & 1.96G & 76.8 \\
 PVT-Tiny & 0.500 & 2.15G & 78.4 \\
 PVT-Tiny & 1.000 & 2.97G & 78.5 \\
        \hline

        \end{tabular}
        \label{tab:retina}
\end{table}

\paragraph{Long ListOps on LRA.}
For all experiments on the LRA benchmark, we follow the released codes of \cite{tay2021long}, implement GRC-Attention using Flax and keep all the other training settings unchanged.
Specifically, all evaluated models are constructed with 512 embedding dimension,  1024 mlp dimension, 8 heads and 6 layers, with only attention functions are replaced by different attention variants and their cached versions.  
Like practice in image classification, GRC modules are initialized with $r=0.5$.
Each model is trained for 5K steps(with 1K steps for warmups) on 2K length sequences individually with batch size 32.
Adam optimizer is adopted with initial learning rates of 0.05 and weight decay of 0.1.

\paragraph{Machine Translation on IWSLT14 and IWSLT15.}
We experiment our methods on widely used public datasets IWSLT14\cite{cettolo-etal-2014-report} and IWSLT15\cite{cettolo-etal-2015-iwslt}.  
For each dataset, we choose three tracks to validate the proposed GRC-Attention, including  German-English(De-En), Spanish-English(Es-En) and  English-French(En-Fr) in IWSLT14 and German-English(De-En), English-Vietnamese(En-Vi) and Czech-English(Cs-En) in IWSLT15.
The Pre-Norm Transformer in \cite{wang2019learning} is used as baselines and the models are implemented using \textit{fairseq-py} \cite{ott2019fairseq} framework.
Following \cite{wang2019learning, ott2019fairseq}, we generally increase the learning rates by 2 and average the last 10 checkpoints for inference. 
The GRC-cached models are derived by replacing their attention functions in Transformer encoders with GRC-Attention modules, which is initialized with caching ratio $r=0.5$ and cache length $T_m=64$.
All the transformers in this task consist of 6 encoder layers and 6 decoder layers,  trained with max length 512 and Adam optimizer.
The  learning rates is initially  0.0015 and then decreased by inverse square root scheduler\cite{ ott2019fairseq}. 

\section{Extensive Results and Ablations}

\subsection{Extensive Results on ImageNet-22k }
We also follow the Swin-Transformer implementation to pre-train our cached Swin-T model on ImageNet-22k. 
 As shown in Tab.~\ref{tab:im22}, GRC effectively enhances the performance of the Swin-T model pre-trained on a larger dataset. In our final version, we'll include additional results of other ViT / Swin-Transformer variants on ImageNet22k.
 
\subsection{Extensive Results on Object Detection }

We extensively apply GRC-Attention to RetinaNet\cite{lin2017focal}, another representative dense prediction network for object detection. 
We choose PVTs\cite{wang2021pyramid} with varying sizes as the backbones, including PVT-Tiny, PVT-Small, and PVT-Medium. 
Like the practice for Mask R-CNN, we use pre-trained PVTs cached by GRC-Attention to initialize the backbone of RetinaNet and train the models for 12 epochs (RetinaNet $1\times$ schedule) with batch size 16 on 8 GPUs.
Following practice in \cite{wang2021pyramid}, we adopt AdamW\cite{loshchilov2018decoupled} optimizer with initial learning rate  $1 \times 10^{-4}$ to update the parameters.
The standard COCO metric Average Precision(AP) is used to evaluate the models. 
Tab.~\ref{tab:retina} shows the detection results using RetinaNet. Consistent to Mask R-CNN, cached PVTs markedly improve their baselines in terms of precision. 
For example, the cached PVT-Medium can achieve 3.8 AP higher than the vanilla PVT-Medium, which is quite significant for this task. 
To sum up, these experiments on downstream tasks (object detection and instance segmentation) demonstrate the generalization capability of the proposed GRC-Attention mechanism in dense vision tasks.

\subsection{Selection of Caching Ratio}
 For main hyper-parameters like caching ratio and memory length, we conduct a series of preliminary experiments and choose the proper ones to achieve better complexity-performance trade-off.  Tab.~\ref{tab:retina} provides ablations towards caching ratios on ImageNet. As shown, we can observe that the performance improvements from larger caching ratio($r$) become marginal when $r > 0.5 $.


\subsection{Training and Inference Throughput}
We compare the throughput(images/sec, per GPU) of GRC-cached models and baselines on ImageNet and the results are shown in Tab.~\ref{tab:th}.  The GPU model is Tesla V100. We can see that GRC improves the performance of PVT models of different sizes while introducing a marginal computational cost increase. Specifically, GRC-cached models surpass their corresponding no-cache baselines by 1.8\%-3.3\% top-1 accuracy with about 15\%-20\% drops in throughput.  
Please also kindly note that although GRC improves model performances with slightly reduced speed, it’s still significantly more efficient than improving models by increasing model depth/width. For example, GRC-cached PVT-Small achieves 81.8\% training accuracy with 146 training throughput, even outperforming the no-cache PVT-Medium which yields 81.2\% accuracy with 101 training throughput.


\end{document}